\documentclass[11pt]{article}
\usepackage{authblk}
\usepackage{eacl2017}
\usepackage{times}
\usepackage{url}
\usepackage{latexsym}
\usepackage{caption}
\usepackage{scalefnt}
\usepackage{covington}
\usepackage{color}
\usepackage{enumitem}
\usepackage{graphicx}
\usepackage{multirow}
\usepackage{amsmath}
\usepackage{amssymb}
\usepackage{latexsym}

\eaclfinalcopy

\makeatletter
\renewcommand*{\p@section}{\S} 
\renewcommand*{\p@subsection}{\S}
\renewcommand*{\p@subsubsection}{\S}
\makeatother

\DeclareSymbolFont{extraup}{U}{zavm}{m}{n}
\DeclareMathSymbol{\spadesuit}{\mathalpha}{extraup}{86}
\DeclareMathSymbol{\vardiamond}{\mathalpha}{extraup}{87}

\title{Joining Hands: Exploiting Monolingual Treebanks for Parsing of Code-mixing Data}

\author{\bf Irshad Ahmad Bhat}
\author{\bf Riyaz Ahmad Bhat}
\author{\bf Manish Shrivastava}
\author{\\ \bf Dipti Misra Sharma}
\affil{LTRC, IIIT-H, Hyderabad, India}
\affil{\tt \{irshad.bhat,riyaz.bhat\}@iiit.ac.in}
\affil{\tt \{m.shrivastava,dipti\}@iiit.ac.in}

\date{}

\begin{document}
\maketitle
\begin{abstract}
In this paper, we propose efficient and less resource-intensive strategies for parsing of code-mixed data. These strategies are not constrained by in-domain annotations, rather they leverage pre-existing monolingual annotated resources for training. We show that these methods can produce significantly better results as compared to an informed baseline. Besides, we also present a data set of 450 Hindi and English code-mixed tweets of Hindi multilingual speakers for evaluation. The data set is manually annotated with Universal Dependencies.
\end{abstract}

\section{Introduction}
Code-switching or code-mixing is a sociolinguistic phenomenon, where multilingual speakers switch back and forth between two or more common languages or language varieties in a single utterance\footnote{For brevity, we will not differentiate between intra- and inter-sentential mixing of languages and use the terms code-mixing and code-switching interchangeably throughout the paper.}. The phenomenon is mostly prevalent in spoken language and in informal settings on social media such as in news groups, blogs, chat forums etc. Computational modeling of code-mixed data, particularly from social media, is presumed to be more challenging than monolingual data due to various factors. The main contributing factors are non-adherence to a standard grammar, spelling variations and/or back-transliteration. It has been generally observed that traditional NLP techniques perform miserably when processing code-mixed language data \cite{solorio2008part,vyas2014pos,ccetinouglu-schulz-vu:2016:W16-58}. 

More recently, there has been a surge in studies concerning code-mixed data from social media \cite[and others]{solorio2008learning,solorio2008learning,vyas2014pos,sharma-EtAl:2016:N16-1,rudra-EtAl:2016:EMNLP2016,joshi-EtAl:2016:COLING}. Besides these individual research articles, a series of shared-tasks and workshops on preprocessing and shallow syntactic analysis of code-mixed data have also been conducted at multiple venues such as Empirical Methods in NLP (EMNLP 2014 and 2016), International Conference on NLP (ICON 2015 and 2016) and Forum for Information Retrieval Evaluation (FIRE 2015 and 2016). Most of these works are an attempt to address preprocessing issues--such as language identification and transliteration--that any higher NLP application may face in processing such data.

Due to paucity of annotated resources in code-mixed genre, the performance of monolingual parsing models is yet to be evaluated on code-mixed structures. This paper serves to fill this gap by presenting an evaluation set annotated with dependency structures. Besides, we also propose different parsing strategies that exploit nothing but the pre-existing annotated monolingual data. We show that by making trivial adaptations, monolingual parsing models can effectively parse code-mixed data.

\section{Parsing Strategies}
We explore three different parsing strategies to parse code-mixed data and evaluate their performance on a manually annotated evaluation set. 
These strategies are distinguished by the way they use pre-existing treebanks for parsing code-mixed data.

\begin{itemize}[leftmargin=*]
\setlength\itemsep{-0.2em}

\item \textbf{Monolingual:}~The monolingual method uses two separate models trained from the respective monolingual treebanks of the languages which are present in the code-mixed data. We can use the monolingual models in two different ways. Firstly, we can parse each code-mixed sentence by intelligently choosing the monolingual model based on the matrix language of the sentence.\footnote{In any code-mixed utterance, the matrix language defines the overall grammatical structure of an utterance, while subordinate language represents any individual words or phrases embedded in the matrix language. We use a simple count-based approach to identify the matrix and subordinate languages of a code-mixed sentence.} A clear disadvantage of this method is that the monolingual parser may not accurately parse those fragments of a sentence which belong to a language unknown to the model. Therefore, we consider this as the baseline method. Secondly, we can linearly interpolate the predictions of both monolingual models at the inference time. The interpolation weights are chosen based on the matrix language of each parsing configuration. The interpolated oracle output is defined as:

\vspace{-2em}
\begin{equation}\label{eq:1}
\begin{split}
y=\operatorname {argmax}(\lambda_m * f(\phi(c_m)) + \\ 
(1-\lambda_m) * f(\phi(c_s)))
\end{split}
\end{equation}
\vspace{-1.5em}

where $f(\cdot)$ is a \textit{softmax} layer of our neural parsing model, $\phi(c_m)$ and $\phi(c_s)$ are the feature functions of the matrix and subordinate languages respectively and $\lambda_m$ is the interpolation weight for the matrix language (see Section \ref{sec:expsetup} for more details on the parsing model).

Instead of selecting the matrix language at sentence level, we define the matrix language individually for each parsing configuration. We define the matrix language of a configuration based on the language tags of top 2 nodes in the stack and buffer belonging to certain syntactic categories such as adposition, auxiliary, particle and verb.

\item \textbf{Multilingual:}~In the second approach, we train a single model on a combined treebank of the languages represented in the code-mixed data. This method has a clear advantage over the baseline Monolingual method in that it would be aware of the grammars of both languages of the code-mixed data. However, it may not be able to properly connect the fragments of two languages as the model lacks evidence for such mixed structures in the augmented data. This would particularly happen if the code-mixed languages are typologically diverse.

Moreover, training a parsing model on augmented data with more diverse structures will worsen the structural ambiguity problem. But we can easily circumvent this problem by including token-level language tag as an additional feature in the parsing model \cite{TACL892}.

\item \textbf{Multipass:}~In the Multipass method, we train two separate models like the Monolingual method. However, we apply these models on the code-mixed data differently. Unlike Monolingual method, we use both models simultaneously for each sentence and pass the input to the models twice. There are two possible ways to accomplish this. We can first parse all the fragments of each language using their respective parsing models one by one and then the root nodes of the parsed fragments would be parsed by the matrix language parsing model. Or, we can parse the subordinate language first and then parse the root of the subordinate fragments with the fragments of matrix language using the matrix language parser. In both cases, monolingual parsers would not be affected by the cross language structures. More importantly, matrix language parser in the second pass would be unaffected by the internal structure of the subordinate language fragments. But there is a caveat, we need to identify the code-mixed fragments accurately, which is a non-trivial task. In this paper, we use token-level language information to segment tweets into subordinate or matrix language fragments.
\end{itemize}

\section{Code-mixed Dependency Annotations}
To the best of our knowledge, there is no available code-mixed data set that contains dependency annotations. There are, however, a few available code-mixed data sets that provide annotations related to language of a token, its POS and chunk tags. For an intrinsic evaluation of our parsing models on code-mixed texts, we manually annotated a data set of Hindi-English code-mixed tweets with dependency structures. The code-mixed tweets were sampled from a large set of tweets of Indian language users that we crawled from Twitter using Tweepy\footnote{http://www.tweepy.org/}--a Twitter API wrapper. We used a language identification system (see \ref{sec:preproc}) to filter Hindi-English code-mixed tweets from the crawled Twitter data. Only those tweets were selected that satisfied a minimum ratio of 30:70(\%) code-mixing. From this data set, we manually selected 450 tweets for annotation. The selected tweets are thoroughly checked for code-mixing ratio. While calculating the code-mixing ratio, we do not consider borrowings from English as an instance of code-mixing. For POS tagging and dependency annotation, we used Universal dependency guidelines \cite{de2014universal}, while language tags are assigned based on the tagset defined in \cite{codeswitch,jamatia2015part}. The annotations are split into testing and tuning sets for evaluation and tuning of our models. The tuning set consists of 225 tweets (3,467 tokens) with a \textit{mixing ratio} of 0.54 and the testing set contains 225 tweets (3,322 tokens) with a \textit{mixing ratio} of 0.53. Here \textit{mixing ratio} is defined as:

\vspace{-1em}
\begin{equation}
 \frac {1}{n}\sum _{s=1}^{n}\frac {H_s}{H_s+E_s}
\end{equation}

where $n$ is the number of sentences in the data set, $H_s$ and $E_s$ are the number of Hindi words and English words in sentence $s$ respectively.

\section{Preprocessing}
\label{sec:preproc}
The parsing strategies that we discussed above for code-mixed texts heavily rely on language identification of individual tokens. Besides we also need normalization of non-standard word forms prevalent in code-mixed social media content and back-transliteration of Romanized Hindi words. Here we discuss both preprocessing steps in brief.

\vspace{-0.5em}
\paragraph{Language Identification}\enspace We model language identification as a classification problem where each token needs to be classified into one of the following tags: `Hindi' (hi), `English' (en), `Acronym' (acro), `Named Entity' (ne) and `Universal' (univ). For this task, we use the feed-forward neural network architecture of Bhat et al. \shortcite{bhatcode}\footnote{Due to space limitation we don't discuss the system architecture in detail. The interested reader can refer to the original paper for a detailed description.} proposed for Named Entity extraction in code mixed-data of Indian languages. We train the network with similar feature representations on the data set provided in ICON 2015\footnote{http://ltrc.iiit.ac.in/icon2015/} shared task on language identification. The data set contains 728 Facebook comments annotated with the five language tags noted above. We evaluated the predictions of our identification system against the gold language tags in our code-mixed development set and test set. Even though the model is trained on a very small data set, its prediction accuracy is still above 96\% for both the development set and the test set. The results are shown in Table \ref{tbl:lidResults}.

\vspace{-0.5em}
\paragraph{Normalization and Transliteration}\enspace We model the problem of both normalization and back-transliteration of (noisy) Romanized Hindi words as a single transliteration problem. Our goal is to learn a mapping for both standard and non-standard Romanized Hindi word forms to their respective standard forms in Devanagari. For this purpose, we use the structured perceptron of Collins \cite{structperceptron} which optimizes a given loss function over the entire observation sequence. For training the model, we use the transliteration pairs (87,520) from the Libindic transliteration project\footnote{https://github.com/libindic/indic-trans} and Brahmi-Net \cite{kunchukuttan2015brahmi} and augmented them with noisy transliteration pairs (63,554) which are synthetically generated by dropping non-initial vowels and replacing consonants based on their phonological proximity. We use Giza++ \cite{och2003systematic} to character align the transliteration pairs for training. 

At inference time, our transliteration model would predict the most likely word form for each input word. However, the single-best output from the model may not always be the best option considering an overall sentential context. Contracted word forms in social media content are quite often ambiguous and can represent different standard word forms such as `pt' may refer to `put', `pit', `pat', `pot' and `pet'. To resolve this ambiguity, we extract n-best transliterations from the transliteration model using beam-search decoding. The best word sequence is then decoded using an exact search over $b^n$ word sequences\footnote{$b$ is the size of beam-width and $n$ is the sentence length. For each word, we extract five best transliterations or normalizations i.e., $b$=5.} scored by a tri-gram language model. The language model is trained on monolingual data using IRSTLM-Toolkit \cite{federico2008irstlm} with Kneser-Ney smoothing. For English, we use a similar model for normalization which we trained on the noisy word forms (3,90,000) synthetically generated from the English vocabulary.

\vspace{0.5em}
\noindent
\begin{minipage}{\linewidth}
\begin{center}
\resizebox{\linewidth}{!}{%
\begin{tabular}{|c|c|c|c|c||c|c|c|c|c|} \hline
\multirow{2}{*}{\textbf{Label}} & \multicolumn{4}{|c||}{\textbf{Development-Set}} & \multicolumn{4}{|c|}{\textbf{Test-Set}} \\ \cline{2-9}
 &  \textbf{Precision} &   \textbf{Recall} & \textbf{F1-Score}  & \textbf{Count}  &  \textbf{Precision} &   \textbf{Recall} & \textbf{F1-Score}  & \textbf{Count} \\ \hline
                                                                                               
 \textbf{acro} &    0.920  &  0.742  &  0.821   &     31    &   0.955   & 0.724   & 0.824    &    29 \\ 
   \textbf{en} &    0.962  &  0.983  &  0.972   &   1303    &   0.952   & 0.981   & 0.966    &  1290 \\ 
   \textbf{hi} &    0.971  &  0.975  &  0.973   &   1545    &   0.968   & 0.964   & 0.966    &  1460 \\ 
   \textbf{ne} &    0.915  &  0.701  &  0.794   &    154    &   0.889   & 0.719   & 0.795    &   167 \\ 
 \textbf{univ} &    0.982  &  0.995  &  0.989   &    434    &   0.987   & 1.000   & 0.993    &   376 \\ \hline \hline
                                                                                               
\textbf{Accuracy} & \multicolumn{3}{|c|}{0.967}   &   3467    & \multicolumn{3}{|c|}{0.961}    &  3322 \\ \hline

\end{tabular}
}
\captionsetup{font=scriptsize}
\captionof{table}{{\label{tbl:lidResults}} Language Identification results on code-mixed development set and test set.}
\end{center}
\end{minipage}
\vspace{1.5em}

\section{Experimental Setup}
\label{sec:expsetup}
The parsing experiments reported in this paper are conducted using a non-linear neural network-based transition system which is similar to \cite{chen2014fast}. The models are trained on Universal Dependency Treebanks of Hindi and English released under version 1.4 of Universal Dependencies \cite{11234-1-1827}.

\vspace{-0.5em}
\paragraph{Parsing Models}\enspace Our parsing model is based on transition-based dependency parsing paradigm \cite{nivre2008algorithms}. Particularly, we use an arc-eager transition system \cite{nivre2003efficient}. The arc-eager system defines a set of configurations for a sentence \textit{w$_1$,...,w$_n$}, where each configuration \textit{C} = (\textit{S}, \textit{B}, \textit{A}) consists of a stack \textit{S}, a buffer \textit{B}, and a set of dependency arcs \textit{A}. For each sentence, the parser starts with an initial configuration where \textit{S} = [ROOT], \textit{B} = [\textit{w$_1$},...,\textit{w$_n$}] and \textit{A} = $\emptyset$ and terminates with a configuration \textit{C} if the buffer is empty and the stack contains the ROOT. The parse trees derived from transition sequences are given by \textit{A}. To derive the parse tree, the arc-eager system defines four types of transitions ($t$): 1) Shift, 2) Left-Arc, 3) Right-Arc, and 4) Reduce.

Similar to \cite{chen2014fast}, we use a non-linear neural network to predict the transitions for the parser configurations. The neural network model is the standard feed-forward neural network with a single layer of hidden units. We use 200 hidden units and RelU activation function. The output layer uses softmax function for probabilistic multi-class classification. The model is trained by minimizing cross entropy loss with an $l2$-regularization over the entire training data. We also use mini-batch Adagrad for optimization \cite{duchi2011adaptive} and apply dropout \cite{hinton2012improving}. 

From each parser configuration, we extract features related to the top four nodes in the stack, top four nodes in the buffer and leftmost and rightmost children of the top two nodes in the stack and the leftmost child of the top node in the buffer. 

\vspace{-0.5em}
\paragraph{POS Models}\enspace We train POS tagging models using a similar neural network architecture as discussed above. Unlike \cite{Collobert2011}, we do not learn separate transition parameters. Instead we include the structural features in the input layer of our model with other lexical and non-lexical units. We use second-order structural features, two words to either side of the current word, and last three characters of the current word.

We trained two POS tagging models: \textit{Monolingual} and \textit{Multilingual}. In the Monolingual approach, we divide each code-mixed sentence into contiguous fragments based on the language tags assigned by the language identifier. Words with language tags other than `Hi' and `En' (such as univ, ne and acro) are merged with the preceding fragment. Each fragment is then individually tagged by the monolingual POS taggers trained on their respective monolingual POS data sets. In the Multilingual approach, we train a single model on combined data sets of the languages in the code-mixed data. We concatenate an additional 1x2 vector\footnote{In our experiments we fixed these to be \{-0.25,0.25\} for Hindi and \{0.25,-0.25 \} for English} in the input layer of the neural network representing the language tag of the current word. Table \ref{tbl:posResults} gives the POS tagging accuracies of the two models.

\vspace{0.5em}
\noindent
\begin{minipage}{\linewidth}
\begin{center}
\resizebox{\linewidth}{!}{%
\begin{tabular}{|c|c|c|c|c||c|c|c|} \hline
\multirow{2}{*}{\textbf{Model}} & \multirow{2}{*}{\textbf{LID}} & \multicolumn{3}{|c||}{\textbf{Development-Set}} & \multicolumn{3}{|c|}{\textbf{Test-Set}} \\ \cline{3-8}
 &  & \textbf{HIN} &   \textbf{ENG} & \textbf{Total}  & \textbf{HIN} &   \textbf{ENG} & \textbf{Total}\\ \hline

\multirow{2}{*}{\textbf{Monolingual}}  & G  &  0.849  &  0.903  &  0.873  &  0.832  &  0.889  &  0.860  \\ 
                                       & A  &  0.841  &  0.892  &  0.866  &  0.825  &  0.883  &  0.853  \\ \hline
\multirow{2}{*}{\textbf{Multilingual}} & G  &  0.835  &  0.903  &  0.867  &  0.798  &  0.892  &  0.843  \\
                                       & A  &  0.830  &  0.900  &  0.862  &  0.790  &  0.888  &  0.836  \\ \hline
\end{tabular}
}
\captionsetup{font=scriptsize}
\captionof{table}{{\label{tbl:posResults}} POS Tagging accuracies for monolingual and multilingual models. LID = Language tag, G = Gold LID, A = Auto LID.}
\end{center}
\end{minipage}
\vspace{1em}

\begin{table*}[htb]
\begin{center}
\resizebox{\linewidth}{!}{%
\begin{tabular}{|c|cc|cc|cc|cccc||cc|cc|cc|cccc|}\cline{1-21}
&\multicolumn{10}{|c||}{\textbf{Gold (POS + language tag)}} & \multicolumn{10}{|c|}{\textbf{Auto (POS + language tag)}} \\\cline{2-21}
Data-set&\multicolumn{2}{|c|}{\textbf{Monolingual}} & \multicolumn{2}{c|}{\textbf{Interpolated}} & \multicolumn{2}{c|}{\textbf{Multilingual}} & \multicolumn{2}{c}{\textbf{Multipass$_f$}} & \multicolumn{2}{c||}{\textbf{Multipass$_s$}} &\multicolumn{2}{|c|}{\textbf{Monolingual}} & \multicolumn{2}{c|}{\textbf{Interpolated}} & \multicolumn{2}{c|}{\textbf{Multilingual}} & \multicolumn{2}{c}{\textbf{Multipass$_f$}} & \multicolumn{2}{c|}{\textbf{Multipass$_s$}} \\\cline{2-21}
& UAS & LAS & UAS & LAS & UAS & LAS & UAS & LAS & UAS & LAS & UAS & LAS & UAS & LAS & UAS & LAS & UAS & LAS & UAS & LAS \\ \hline
\multicolumn{1}{|c|}{CM$_d$} & 60.77 & 49.24 & 74.62 & 64.11 & \textbf{75.77} & \textbf{65.32} & 69.37 & 58.83 & 70.23 & 59.64 & 55.80 & 43.36 & \textbf{68.24} & \textbf{56.07} & 67.71 & 55.18 & 63.34 & 52.22 & 64.60 & 53.03 \\
\multicolumn{1}{|c|}{CM$_t$} & 60.05 & 48.52 & \textbf{74.40} & 63.65 & 74.16 & \textbf{64.11} & 68.54 & 57.87 & 69.12 & 58.64 & 54.95 & 43.03 & 65.14 & 54.00 & \textbf{66.18} & \textbf{54.40} & 62.37 & 51.11 & 63.74 & 52.34\\\hline
\multicolumn{1}{|c|}{HIN$_t$} & \textbf{93.29} & \textbf{90.60} & 92.61 & 89.64 & 91.96 & 88.46 & \textbf{93.29} & \textbf{90.60} & \textbf{93.29} & \textbf{90.60} & \textbf{91.92} & \textbf{88.39} & 91.82 & 88.34 & 89.52 & 84.83 & \textbf{91.92} & \textbf{88.39} & \textbf{91.92} & \textbf{88.39}\\
\multicolumn{1}{|c|}{ENG$_t$} & 85.12 & \textbf{82.86} & 84.21 & 81.82 & \textbf{85.16} & 82.79 & 85.12 & \textbf{82.86} & 85.12 & \textbf{82.86} & \textbf{83.28} & \textbf{79.90} & 82.08 & 78.54 & 82.53 & 79.11 & \textbf{83.28} & \textbf{79.90} & \textbf{83.28} & \textbf{79.90}\\\hline
\end{tabular}}%

\captionsetup{font=scriptsize}
\caption{{\label{tbl:parsingResults}} Accuracy of different parsing strategies on Code-mixed as well as Hindi and English evaluation sets. CM$_{d|t}$ = Code-mixed development and testing sets; HIN$_t$ = Hindi test set; ENG$_t$ = English test set; Multipass$_{f|s}$ = fragment-wise and subordinate-first parsing methods.}
\end{center}
\end{table*}

\paragraph{Word Representations}\enspace For both POS tagging and parsing models, we include the lexical features in the input layer of the Neural Network using the pre-trained word representations while for the non-lexical features, we use randomly initialized embeddings within a range of $-0.25$ to $+0.25$.\footnote{Dimensionality of input units in POS and parsing models: 80 for words, 20 for POS tags, 2 for language tags and 20 for affixes.}~We use Hindi and English monolingual corpora to learn the distributed representation of the lexical units. The English monolingual data contains around 280M sentences, while the Hindi data is comparatively smaller and contains around 40M sentences. The word representations are learned using Skip-gram model with negative sampling which is implemented in {\tt word2vec} toolkit \cite{mikolov2013efficient}. For multilingual models, we use robust projection algorithm of Guo et al. \shortcite{guo2015cross} to induce bilingual representations using the monolingual embedding space of English and a bilingual lexicon of Hindi and English ($\sim$63,000 entries). We extracted the bilingual lexicon from ILCI and Bojar Hi-En parallel corpora \cite{jha2010tdil,hindencorp05:lrec:2014}.

\section{Experiments and Results}
We conducted multiple experiments to measure effectiveness of the proposed parsing strategies in both gold and predicted settings. In predicted settings, we use the monolingual POS taggers for all the experiments. We used the Monolingual method as the baseline for evaluating other parsing strategies. The baseline model parses each sentence in the evaluation sets by either using Hindi or English parsing model based on the matrix language of the sentence. For baseline and the Multipass methods, we use bilingual embedding space derived from matrix language embedding space (Hindi or English) to represent lexical nodes in the input layer of our parsing architecture. In the Interpolation method, we use separate monolingual embedding spaces for each model. The interpolation weights are tuned using the development set and the best results are achieved at $\lambda_m$ ranging from 0.7 to 0.8 (see eq. \ref{eq:1}). The results of our experiments are reported in Table \ref{tbl:parsingResults}. Table \ref{tbl:beamResults} shows the impact of sentential decoding for choosing the best normalized and/or back-transliterated tweets on different parsing strategies (see \ref{sec:preproc}).

\vspace{0.5em}
\noindent
\begin{minipage}{\linewidth}
\begin{center}
\resizebox{\linewidth}{!}{%
\resizebox{\linewidth}{!}{
\begin{tabular}{|c|cc|cc||cc|cc|}\hline
 & \multicolumn{4}{|c||}{\textbf{First Best}} & \multicolumn{4}{c|}{\textbf{K-Best}} \\ \cline{2-9}
Data-set & \multicolumn{2}{|c|}{\textbf{Multilingual}} & \multicolumn{2}{c||}{\textbf{Interpolated}} & \multicolumn{2}{c|}{\textbf{Multilingual}} & \multicolumn{2}{c|}{\textbf{Interpolated}} \\\cline{2-9}
         & UAS   & LAS   & UAS   & LAS      & UAS   & LAS   & UAS   & LAS     \\ \hline
CM$_d$   & 66.21 & 53.55 & 66.70 & 53.68    & 67.71 & 55.18 & 68.24 & 56.07 \\
CM$_t$   & 65.87 & 53.92 & 64.26 & 53.35    & 66.18 & 54.40 & 65.14 & 54.00 \\\hline
\end{tabular}}
}
\captionsetup{font=scriptsize}
\captionof{table}{{\label{tbl:beamResults}} Parsing accuracies with exact search and k-best search (k = 5). CM$_{d|t}$ = Code-mixed development and testing sets.}
\end{center}
\end{minipage}
\vspace{1em}

\vspace{1em}
All of our parsing models produce results that are at-least 10 LAS points better than our baseline parsers which otherwise provide competitive results on Hindi and English evaluation sets \cite{straka2016udpipe}.\footnote{Our results are not directly comparable to \cite{straka2016udpipe} due to different parsing architectures. While we use a simple greedy, projective transition system, Straka et al. \shortcite{straka2016udpipe} use a search-based swap system.} Among all the parsing strategies, the Interpolated methods perform comparatively better on both monolingual and code-mixed evaluation sets. Interpolation method manipulates the parameters of both languages quite intelligently at each parsing configuration. Despite being quite accurate on code-mixed evaluation sets, the Multilingual model is less accurate in single language scenario. Also the Multilingual model performs worse for Hindi since its lexical representation is derived from English embedding space. It is at-least 2 LAS points worse than the Interpolated and the Multipass methods. However, unlike the latter methods, the Multilingual models do not have a run-time and computational overhead. In comparison to Interpolated and Multilingual methods, Multipass methods are mostly affected by the errors in language identification. Quite often these errors lead to wrong segmentation of code-mixed fragments which adversely alter their internal structure. 

Despite higher gains over the baseline models, the performance of our models is nowhere near the performance of monolingual parsers on newswire texts. This is due to inherent complexities of code-mixed social media content \cite{solorio2008part,vyas2014pos,ccetinouglu-schulz-vu:2016:W16-58}.

\section{Conclusion}
In this paper, we have evaluated different strategies for parsing code-mixed data that only leverage monolingual annotated data. We have shown that code-mixed texts can be efficiently parsed by the monolingual parsing models if they are intelligently manipulated. Against an informed monolingual baseline, our parsing strategies are at-least 10 LAS points better. Among different strategies that we proposed, Multilingual and Interpolation methods are two competitive methods for parsing code-mixed data. 

The code of the parsing models is available at the GitHub repository\enspace\path{https://github.com/irshadbhat/cm-parser}, while the data can be found under the Universal Dependencies of Hindi at\enspace\path{https://github.com/UniversalDependencies/UD_Hindi}.

\vspace{-1em}
\bibliographystyle{eacl2017}
\bibliography{eacl2017}

\end{document}